\documentclass{article}

\usepackage[final]{amlc_2023}

\usepackage[utf8]{inputenc} % allow utf-8 input
\usepackage[T1]{fontenc}    % use 8-bit T1 fonts
\usepackage[colorlinks,linkcolor=red,anchorcolor=blue,citecolor=blue]{hyperref}
\usepackage{natbib}
\usepackage{url}            % simple URL typesetting
\usepackage{booktabs}       % professional-quality tables
\usepackage{amsfonts}       % blackboard math symbols
\usepackage{nicefrac}       % compact symbols for 1/2, etc.
\usepackage{microtype}      % microtypography
\usepackage[dvipsnames]{xcolor}         % colors
\usepackage{soul}

\usepackage{algpseudocode}
\usepackage{amsmath}
\usepackage{cleveref}
\usepackage{bm}
\usepackage{amssymb}
\usepackage{algorithm}
\usepackage{bbm}
\usepackage{tikz}
\usetikzlibrary{shapes.geometric, arrows}
\usepackage{graphicx}
\usepackage{subcaption}
\usepackage{tabularx}
\usepackage{multirow}
\usepackage{lineno}

\usepackage{lipsum}
\newcommand\blfootnote[1]{%
  \begingroup
  \renewcommand\thefootnote{}\footnote{#1}%
  \addtocounter{footnote}{-1}%
  \endgroup
}

\newcommand{\B}{\textbf}

\title{Deep RL Dual Sourcing Inventory Management with Supply and Capacity Risk Awareness}

\author{%
  Defeng Liu
  \\ \texttt{liudef@amazon.com} \\
  \And
  Ying Liu \\
   \texttt{liuyingy@amazon.com} \\
 \And
 Carson Eisenach \\
    \texttt{ceisen@amazon.com} \\
}

\begin{document}

\maketitle

\begin{abstract}
In this work, we study how to efficiently apply reinforcement learning (RL) for solving large-scale stochastic optimization problems by leveraging intervention models. The key of the proposed methodology is to better explore the solution space by simulating and composing the stochastic processes using pre-trained  deep learning (DL) models. We demonstrate our approach on a challenging real-world application, the multi-sourcing multi-period inventory management problem in supply chain optimization. In particular, we employ deep RL models for learning and forecasting the stochastic supply chain processes under a range of assumptions. Moreover, we also introduce a constraint coordination mechanism, designed to forecast dual costs given the cross-products constraints in the inventory network. We highlight that instead of directly modeling the complex physical constraints into the RL optimization problem and solving the stochastic problem as a whole, our approach breaks down those supply chain processes into scalable and composable DL modules, leading to improved performance on large real-world datasets. We also outline open problems for future research to further investigate the efficacy of such models.
\end{abstract}

\blfootnote{\textit{Accepted by the
$\mathit{42}^{nd}$ International Conference on Machine Learning, Workshop on Scaling Up Intervention Models}, 
Vancouver, Canada. PMLR 267, 2025.
Copyright 2025 by the author(s).}

\section{Introduction}
Modern inventory management systems (IMS) in retail supply chain (such as Walmart and Amazon) typically deploy multi-sourcing buying strategies, which employ a combined system with a just-in-time (JIT) ordering strategy plus other specialized strategies that strive to achieve a balance between supply shortage and inventory health across all products without compromising the inventory management's contribution to the overall retail service.

In real world supply chains, order quantities are subject to several post-processors, including modification to meet vendor constraints such as minimum order and batch size constraints. Second, the supply may be unreliable and vendors may only partially fill orders that they receive. This may occur for multiple reasons, including that the vendor itself is out of stock. In the literature the proportion of the original order quantity retailer ultimately receives is referred to as the yield or fill rate. In the current state, there is no existing representation model to learn those external processes.

On the other hand, capacity limitations, both in terms of network inbound capacity and storage capacity, require the supply chain system to assess which inventory to purchase and when to have it arrive through all of its buying channels in order to have an effective and efficient capacity controller (CC). For example, for JIT buys, traditional CC mechanisms typically have the ability to simulate and compute weekly capacity costs to meet the available capacity for short forward-looking periods, e.g. one to three months. However, as modern sourcing systems have changed to reflect increasing complexity in global supply chain and the availability of direct-from-manufacturer cost discounts,  inventory decisions are made further into the future to cover planning horizons of longer periods, e.g. more than six months.
% These penalties are dynamically adapted every week based on the changing demand forecast, inventory picture and the ASIN economics. 

This work is also motivated by prior works~\citep{madeka2022deep, andaz2024learninginventorycontrolpolicy, eisenach2024neuralcoordinationcapacitycontrol} which established the viability of Deep RL for single-sourcing inventory planning problem. But we forward this line of research by targeting for a more complex variant of the aforementioned problem, wherein multi-sourcing channels of vendors are available and the primary trade-off considered was to strike a balance for costs vs supply risks for different vendors. On the other hand, we aim to investigate effects of discount rates on long lead orders, stochastic quantity over time arrival profiles.

Here, we emphasize that a critical issue encountered in solving the multi-sourcing inventory problem via traditional methods such as dynamic programming is the unknown dynamics of a variety of underlying processes associated with inventory control. For instance, customer demand is not deterministic and exhibits volatility which are influenced by seasonality, external sourcing processes, etc. To incorporate those complex supply chain processes, we attempt to investigate whether it is possible to efficiently scale the training of decision policies from the vast amount of available historical data, without the requirement to build separate models for the various state variables with unknown {stochastic behaviors}.

We organize rest of the paper as summarized next. Section \ref{sec: related_work} reviews the related work.  In Section \ref{sec: problem_formln}, we mathematically formulate the dual-sourcing inventory management problem and thereby describe our dual sourcing RL methodology as a solution for the problem. In Section \ref{sec: evals_plan}, we present our main results from numerical experiments. Additionally, in Section \ref{sec: capacity}, we extend our dual sourcing RL baseline model by introducing a capacity mechanism control strategy and provide evaluation results. Finally, we conclude the paper with discussion on future work in Section \ref{sec: conclusion}. 

\section{Related work}
\label{sec: related_work}

\noindent \textbf{Dual Sourcing Inventory Control:} Dual sourcing is a heavily studied problem in inventory control literature, and has become a common practice supply chain organizations worldwide. In vanilla dual sourcing problem, the sourcing channels i.e., JIT, LLT essentially cause a trade-off in lead time vs ordering costs since usually LLT ( direct import sourcing channel ) has lower sourcing cost. In this context, Tailored Base-Surge (TBS) policy is a widely used method, wherein at each period, a constant order is placed via the LLT channel, and a dynamic order is placed to match an inventory order-up-to level via the JIT channel respectively. The JIT channel's order-up-to level essentially implies a \textit{Safety Stock} maintained to address sudden \textit{demand surges} \citep{xin2018asymptotic}. Although, such policies present fairly intuitive approaches to handle the dual sourcing problem, but the optimality analysis of such policies has been a hard problem in general \citep{whittemore1977optimal}, except for certain edge cases. Interestingly, the optimal policy is shown to be vanilla base-stock policy when the lead time difference between LLT and JIT channels is exactly 1 period \citep{fukuda1964optimal}.

\noindent \textbf{Reinforcement Learning for Supply Chain Optimization} Recently, the progress of applying Machine Learning in various optimization problems \citep{bello2016neural, silver2016mastering,khalil2017learning,nazari2018reinforcement,liu2021learning, bengio2021machine, Liu_Fischetti_Lodi_2022, liu2022machine, sinclair2023hindsight,liu2023machine, shao2024deepseekmath} motivates its application in inventory management problems \citep{madeka2022deep,qi2023practical,liang2024reinforcement,liang2024dual,chen2025learning}. In this context, Deep RL approaches have emerged more recently over its other \textit{Model-Based} counterparts, due to improved computational scalability and generalization performance of Deep Neural Network (DNN) architectures. Furthermore, DRL methods have been shown to achieve performance gains over benchmarks for the multi-period newsvendor problem under fairly realistic assumptions on costs, prices, demand and stationarity \citep{balaji2019orl}. It is worth highlighting that DRL for single sourcing problem has been comprehensively investigated with extensive empirical evidences at Amazon \citep{madeka2022deep}. 

\noindent \textbf{Capacity Management and Coordinatation Mechanism}
Retailers typically manage a supply chain for multiple products and has limited capacity resources (such as storage) that are shared amongst all the products that retailer stocks. The classic method for handling capacity constraints is to call a simulation and optimization process to compute shadow prices on the shared resources. Model predictive control (MPC) consists of using a model to forward simulate a system to optimize control inputs and satisfy any constraints. At each time step, one re-plans based on updated information that has become available in order to select the next control input. 
Recent work \citet{maggiar2024consensus} introduced the Consensus Planning Protocol (CPP), which targets problems where multiple agents (each of which is locally optimizing its own utility function) all consume a shared resource. This is closely related to a distributed ADMM procedure \citep{boyd2011distributed}. Another work \cite{eisenach2024neuralcoordinationcapacitycontrol} presented a new capacity control mechanism for RL-based buying policies and proposed a Neural Coordinator model to generate forecasts of capacity prices. Their formulation of capacitated inventory management can be viewed as a special case of CPP (a central coordinator adjusts prices, and the other agents adjust their plans).

\section{A Deep RL Approach for the Dual Sourcing Problem}
\label{sec: problem_formln}

% The problem formulation below broadly follows the Dual Sourcing Problem from~\cite{feng2023deep}. To keep the document self-contained, we reproduce the important details in Section \ref{sec: problem_formln}. However, those familiar with the existing formulation may skip to Section \ref{sec: evals_plan} and Section \ref{sec: capacity}.

% \subsection{Mathematical Notations}
% \textcolor{green}{can you add a diagram to show the time stamp of planning and targeting? here is an example}
%  https://drive.corp.amazon.com/documents/maggiara@/Shared_Files/wiki_page_docs/Projects/2018/LSTB_toys%20-%20RSS.pdf
% \textcolor{red}{when you explain the denotations, try to match with this paper, for example, new inventory procurement costs is actually product cost}
% https://drive.corp.amazon.com/documents/maggiara@/Shared_Files/wiki_page_docs/Documentation/Buying_101.pdf

\subsection{Modeling the Dual Sourcing Problem as an Exo-IDP}

In this section we model the dual sourcing problem as an \textit{Exogenous Interactive Decision Process} (Exo-IDP) following the framework introduced in \citet{madeka2022deep}. In an Exo-IDP, the state describing the system dynamics decompose into (i) {\em Exogenous} processes, which evolve independently of the buy policy's actions, and (ii) {\em Endogenous} processes that are impacted by the agent's actions and the exogenous signals. 

We consider the case of a retailer managing a set $\mathcal{A}$ of products for $T$ time steps, where the objective is to maximize revenue by placing orders to both \textit{long lead time} (LLT) and JIT sources. To succinctly describe our process, we focus on just one product $i \in \mathcal{A}$, though we note that decisions can be made jointly for every product.

% $\epsilon_{t}^{i}$ on the cost of goods sold $\tilde{c}_{t}^{i}$
\paragraph{State} The price received at sale, costs incurred on JIT purchase, and holding costs are denoted as $p_{t}^{i}$, $c_{t}^{J,i}$, and $h_{t}^{i}$, respectively. For LLT orders, the retailer typically receives a discount on the cost of goods sold, so there is a different cost incurred on purchase $c_t^{L,i}$. , Additionally, the demand for product $i$ at time $t$ is denoted as $d_{t}^{i}$. The aforementioned set of variables are completely exogenous, and therefore their evolution is independent of any policy's interaction with the Exo-IDP.

Together these exogenous processes form the state as follows,
\begin{align}
    s_{t}^{i} \triangleq \big( d_{t}^{i}, p_{t}^{i}, c_{t}^{J,i}, c_{t}^{L,i}, , \boldsymbol{\rho}^{J,i}_{t}, \boldsymbol{\rho}^{L,i}_{t}, \boldsymbol{M}^{J,i}_t, \boldsymbol{M}^{L,i}_t\big).
\end{align}

The history of the joint process up to time $t$ is defined as
\[
H_t := \{(k_0^i, s_1^i,\dots,s_t^i) \}_{i=1}^{|\mathcal{A}|},
\]
where $k_0^i$ is the initial inventory level. Product-level histories can be defined similarly.

\paragraph{Actions} For product $i$, action $a_{t}^{i}$ implies placing orders via JIT, LLT channels at time $t$. In other words, agent's interaction with the Exo-IDP is only via placing orders. More precisely,
\begin{align}
    a_{t}^{i} \triangleq (q^{J,i}_{t}, q^{L,i}_{t}).
\end{align}
So we have $a_t^i \in \mathbb{R}^{2}$. For a class of policies parameterized by $\theta$, we can defined the actions as
\begin{align}
    a_t^i = \pi^i_{\theta, t}(H_t). \label{eq: action_vec_def2}   
\end{align}

We define the set of these policies as $\bm{\Pi} \triangleq \{\pi_{t}^{i}(; \theta) | \theta \in \Theta, i \in \mathbb{A}, t \in [0, T-1] \}$.

% It is important to note that the policy class defined above is very general and in an actual implementation of the RL agent, one would include features such as current inventory levels -- this fits into the framework described here because {\it the inventory is itself a function of only $H_t$ and the policy parameters $\theta$}. For detailed feature list used by our policies, see in Appendix \ref{sec: features}. In our implementation, we also configure the history vector $\mathcal{H}_t$ with a fixed length from period $(t-m)$ till $t$, where $m$ is a pre-determined length parameter. 

\paragraph{External Sourcing Processes}
 After orders are created and submitted to vendors, they can arrive in multiple shipments over time, and the total arriving quantity may not necessarily sum up to the order quantity placed. Any constraints the vendor imposes on the retailer's orders $\boldsymbol{M}^{J,i}_t \in \mathbb{R}^{d_v}$ -- such as minimum order quantities and batch sizes -- are exogenous to the ordering decisions. We define an order quantity post-processor on the JIT source $f^J_{p} : \mathbb{R}_{\geq 0} \times \mathbb{R}^{d_v} \rightarrow \mathbb{R}_{\geq 0}$ that may modify the order quantity. The final order quantity submitted to the vendor is denoted as $\tilde{q}^{J,i}_t := f^J_{p}(q_t^{J,i}, \mathrm{M}^{J,i}_t)$. Similarly, the final LLT order quantity is defined as $\tilde{q}^{L,i}_t := f^L_{p}(q_t^{L,i}, \mathrm{M}^{L,i}_t)$.
 
 At every time $t$, the vendor has allocated a supply $U^{J,i}_t$ that denotes the maximum number of units it can send (regardless the amount we order), which will arrive over from the current week up to $L_1$ weeks in the future according to an exogenous {\it arrival shares} process $(\rho^{J,i}_{t,0},...,\rho^{J,i}_{t,L_1})$ where $\sum_l \rho^{J,i}_{t,l} = 1$ and $\rho^{J,i}_{t,l} \geq 0$ for all $i$, $t$ and $l$. 
 % Similarly we can define the arrivals process of the long-lead-time (LLT) source as arriving in at most $L_2$ weeks, with its own arrival shares process $(\rho^{L,i}_{t,0},...,\rho^{L,i}_{t,L_2})$, supply process $U^{L,i}_t$, and vendor constraints $\boldsymbol{M}^{L,i}_t \in \mathbb{R}^{d_v}$.
 The arrival quantity at lead time $j$ from order $q^{J,i}_t$ can be denote as $o^{J,i}_{t,j} := \min(U^{J,i}_{t}, \tilde{q}^{J,i}_{t})\rho^{J,i}_{t,j}$.  The LLT arrival quantities are defined similarly and denoted as $o^{L,i}_{t,j} := \min(U^{L,i}_{t}, \tilde{q}^{L,i}_{t})\rho^{L,i}_{t,j}$. 

 In brief, the overall sourcing processes from the initial order quantity to final arrivals can be modeled as
 \begin{align}
     o_{t,j}^{J,i} &:= \min(U^{J,i}_t,f_p(q_t^{J,i},\boldsymbol{M}^{J,i}_t))\rho^{J,i}_{t,j}, \\
     o_{t,j}^{L,i} &:= \min(U^{L,i}_t,f_p(q_t^{L,i},\boldsymbol{M}^{L,i}_t))\rho^{L,i}_{t,j}. 
 \end{align}

\paragraph{Internal Inventory Dynamics} 

$I_{t-}^{i}$ and $ I_{t}^{i}$ denote the on-hand inventory for product $i$ at the beginning and end of a period $t$, respectively. The inventory update rule is given as follows,
\begin{equation}
	\label{eq:arrivals}
	I^i_{t_-} = I^i_{t-1} + \sum_{j=0}^{L_1} o^{J,i}_{t,j} + \sum_{j=0}^{L_2} o^{L,i}_{t,j},
\end{equation}
and
\begin{equation}
	\label{eqn:onhand_periodend1}
	I_{t}^{i} = \max\{I_{t-}^{i} - d_{t}^{i}, 0 \}. 
\end{equation}

% \noindent \underline{\textit{Confirmation Rate and Inventory Arrivals:}} We incorporate the idea of inventory arrivals over time via the JIT, LLT channels according to their respective distributions $\{\mathrm{P}_{J}, \mathrm{P}_{L}\}$. 

% For the JIT channel, we have integrated GEN-QOT model \cite{genqot} for $\mathrm{P}_{J}$. For the long lead channel, we have the following assumption over the confirmation rate.

\paragraph{Reward Function} We formulate the reward realized at $t$ taking into account the current period inventory costs and sales. The construction of reward in our problem is such that it measures the periodic cash outflows caused due to replenishment and holding costs of inventory, and inflows are attributed to customer sales. It is defined as
\begin{align}
    \label{eqn:rewardfunc}
    R_{t}^{i} \triangleq p^i_t\min(d_t^i, I^i_{t-}) - c_t^{J,i} \sum_{j=0}^{L_1} o^{J,i}_{t,j} - c_t^{L,i}\sum_{j=0}^{L_1} o^{L,i}_{t,j} - h_{t}^{i}I_{t}^{i}.
\end{align}

Hence, computation of reward is essentially a function of history vector $H_t$ and policy parameters $\theta$, $R(\mathcal{H}_t, s_{t}^{i}, \theta)$.\footnote{Reward is a function of current period action $a_{t}^{i}$ which is computed via policy $\pi$ parameterized by $\theta$, and input $\mathcal{H}_{t}$.}

% \paragraph{Exo-IDP}
% Here we introduce a new notation to express the time value discounting factor for the reward by $\gamma$. It is necessary to emphasize that the two discount parameters i.e., $\epsilon, ~\gamma$, have very different implications in our problem setting. While, $\epsilon$ is the price discount applied on the order quantities procured through long lead channel, $\gamma$ is the discount factor that accounts for time value of for the overall rewards, this controls how much importance does future rewards have. Usually $\gamma \in [0, 1]$, where 0 implies the extreme case where only immediate period rewards are important. 

% Finally, we define the Exo-IDP of the dual sourcing problem as:
% \begin{align}
%     \mathcal{I} \triangleq \langle \mathcal{H}, \mathcal{A}, \gamma, T,  \sum_{t=0}^{T-1} \sum_{i \in \mathbb{A}} R^{i}_{t}, \bm{\Pi} \rangle,
% \end{align}
% where $\mathcal{H}$ is entire history of state evolution.

\subsection{The Dual Sourcing Optimization Problem}
In our setting, we aim to maximize the total discounted \footnote{This discounted reward implies time value of actual cash flows. The discounting factor here is $\gamma$.} reward across the $T$ length time horizon in expectation, while accounting for other constraints.

In the following, we state the dual sourcing optimization problem $\mathbf{\mathcal{P}_1}$,
\begin{align}
     \mathbf{\mathcal{P}_1}: & ~\underset{\theta \in \Theta}{\max} ~\mathbb{E} \Big[\sum_{i \in \mathbb{A}} \sum_{t = 0}^{T-1} \gamma^{t} R_{t}^{i}(\theta) \Big], \label{eq: p_1} \\
    & \text{s.t.}   \nonumber \\
    & I_{0}^{i} = \overline{I}^{i},  \label{eq: inv_init_1} \\
    & \text{Equations} (3-7),
\end{align}
where Eq. \eqref{eq: p_1} is the expression for initial inventory.

\subsection{Scaling the Learning by Forecasting Sourcing Processes}
In practice, we do not actually observe the full supply and arrival share processes for either the JIT or LLT sources. Under the IDP model described in the previous section, we only observe the arrivals share processes when an order was placed historically and we only observe the supply process when we do not receive the full order quantity. To handle this censoring we directly forecast the {\it arrivals} \citep{andaz2024learninginventorycontrolpolicy} instead of the supply and arrival shares processes. 

To see why this makes sense, note that the dynamics \eqref{eq:arrivals} and reward function \eqref{eqn:rewardfunc} depend only on the arrivals $o_{t,j}^{J,i} := \min(U^{J,i}_t,f_p(q_t^{J,i},\boldsymbol{M}^{J,i}_t))\rho^{J,i}_{t,j}$ and $o_{t,j}^{L,i} := \min(U^{L,i}_t,f_p(q_t^{L,i},\boldsymbol{M}^{L,i}_t))\rho^{L,i}_{t,j}$. Thus, for the purposes of constructing our simulator from historic data,  we forecast arrivals conditional on the action $a_t^i$ rather than estimating the post-processing behavior and the supply and arrival share processes.

\paragraph{Arrivals}
For arrivals, denoting by $H_{t,O}^i$ the observed components of $H_t^i$, one can forecast
\begin{equation}
\label{eqn:qot-pred}
	p(o^{J,i}_{t,0},\dots, o^{J,i}_{t,L_1} | H_{t,O}^i; \psi_1).
\end{equation}
Note that conditioning on $ H_{t,O}^i$, $\theta$ is equivalent to conditioning on $ H_{t,O}^i$, $\psi_1$ and the past JIT order actions $a_s^i$ for $s \leq t$. 

% \paragraph{LLT Arrivals} It is also important to note that in the current RL simulator, we only applied the QOT forecasting model for JIT arrivals. For LLT arrivals, we assume $x$ arrival rate with lead time $\delta^{L,i}$, where $0 < \delta^{L,i} < L_2$ and $L_2$ is the maximal lead time. The LLT arrivals are given as follows
% \begin{equation}
% \label{eqn:llt_arrivals}
% 	o^{L,i}_{t,\delta^{L,i}} =  x q_t^{L,i},
% \end{equation}
% and
% \begin{equation}
% \label{eqn:llt_arrivals}
% 	o^{L,i}_{t,k} = 0.
% \end{equation}
% where $k = 0, \dots, \delta^{L,i}-1, \delta^{L,i}+1, \dots, L_2$.

\subsection{The Deep RL Algorithm for Training the Buy Policy}
\label{sec: proposed_method}

Firstly, recall that parameters $\theta \in \Theta$ at any time $t$ essentially dictates the RL agent's policy as expressed in eq. \eqref{eq: action_vec_def2}, therefore our problem reduces to learning optimal parameters $\theta^{\star} \in \Theta$. Specifically, we leverage a Deep Neural Network (DNN) architecture for $\pi(\cdot, \cdot; \theta)$. So, $\theta$ in our DRL framework essentially implies the weights and parameters of the constituent neurons in the policy network. We applied the same DNN architecture as used in \cite{madeka2022deep}. The detailed training algorithm can be found in Appendix \ref{sec_app: training}.

\section{Experimental Evaluations} \label{sec: evals_plan}

\subsection{Training/Evaluation Configurations}
\paragraph{Real-world Dataset} We use the same real-world dataset as used in \cite{madeka2022deep} with approximately 80,000 products for 124 weeks from April 2017 to August 2019. 
Out of the 124 weeks, we treat the first 72 weeks as training dataset and the remaining 52 as the backtesting dataset. However, in future iterations, we plan to train the DRL models on 104 weeks so that the policy agents can better track seasonal patterns.

% \noindent \underline{\textit{Quantity over Time Sampling:}} For the JIT channel, we have integrated GEN-QOT model \cite{genqot} for $\mathbbm{P}_{J}$. For LLT orders, in our experiments we have investigated model performances under $100\%$ confirmation rates and fixed $10\%$ discounts $\epsilon=0.1$. 

% We use Bernoulli distributions as function of lead time for sampling instantaneous LLT confirmation rates $\lambda_{L, t}^{i}$. Whereas, JIT confirmation rates and lead times $\{\lambda_{J, t}^{i}, \xi_{J,t}^{i} \}$ are jointly obtained via GEN-QOT.
%, a brief overview of this sampling model is provided in Appendix \ref{sec: vlt_dist}.\\

% \noindent \underline{Inventory Initialization:} Note that the behavior of the backtested policy can be drastically different with different initialization choices for the starting inventory $I_{0}^{i}$. In our experimental investigations, we initialize the inventory where another policy left off. Such an initialization is often found to be more stable in practice and helps reduce the real world noise.  

% The central idea is that we would want to investigate how different backtested policies will react to initializations where initial inventory is either very starved or has favorable levels.

\paragraph{Baselines}
We compare our Dual Sourcing RL (\texttt{DualSrc-RL}) buy policy against several baselines, i.e. \texttt{BaseStockHorizonTip (BSHT)}, \textit{Tailored Base Surge (TBS)}, \textit{Just-in-time RL (JIT-RL)}. The first two are classic operation research baselines and the \textit{JIT-RL} is the RL baseline where the RL model is trained as a single-sourcing model \citep{madeka2022deep}. More detailed description of those baselines are reported in Appendix \ref{sec: baseline_methods}.

% \paragraph{Inventory Initialization} 
% In our experiments, we test the compared policies on two initialization scenarios, a starved initial inventory and a relatively much higher level of inventory over the two experimental scenarios.

\subsection{Main Results and Analytics}
%\subsection{Brief Overview of Training/Evaluation Configurations}
We use the training dataset to train the RL algorithms and perform evaluation experiments for the compared algorithms over the test dataset. The evaluation results are reported in Table \ref{tab: backtest cs1}.
%Out of the 124 weeks, we treat the first 72 weeks as training dataset and the remaining 52 as the backtesting dataset. However, in future iterations, we plan to train the DRL models on 104 weeks so that the policy agents can better track seasonal patterns. We perform two different set of experiments with either a starved initial inventory or a relatively higher inventory.
%\subsection{Results with different inventory initializations}

\begin{table}[H]
\begin{center}
\begin{tabular}{|c|c|c|}
    \hline
    Setting & Method  & \% of \texttt{BSHT} \\
    \hline
  \multirow{2}{*}{JIT Policy} & \texttt{BSHT}  & 100 \\
    \cline{2-3}
    & \texttt{JIT-RL}  & 104.78 \\
    \hline
    \multirow{2}{*}{Dual Sourcing}& \texttt{TBS}   & 117.69 \\
    \cline{2-3}
    &\texttt{DualSrc-RL}  & 121.54 \\
   
    \hline
\end{tabular}
\end{center}
\caption{Cumulative discounted rewards (as \% of \texttt{BSHT}) for 52 backtest periods for different policy methods.}
\label{tab: backtest cs1}
\end{table}
% \vspace{-10mm}
% \begin{table}[H]
% \begin{center}
% \begin{tabular}{|c|c|c|}
%     \hline
%     Setting & Method  & \% of \texttt{BSHT} \\
%     \hline
%   \multirow{2}{*}{JIT Policy} & \texttt{BSHT}   & 100 \\
%     \cline{2-3}
%     & \texttt{JIT-RL}   & 105.34 \\
%     \hline
%     \multirow{2}{*}{Dual Sourcing}& \texttt{TBS} & 103.64 \\
%     \cline{2-3}
%     &\texttt{DualSrc-RL}   & 108.47 \\
%     \hline
% \end{tabular}
% \end{center}
% \caption{Cumulative discounted rewards (as \% of \texttt{BSHT}) for 52 backtest periods for different policy methods. Experiment Scenario 2: Higher inventory initialization.}
% \label{tab: backtest cs2}
% \end{table}
\vspace{-5mm}
From the results above, we can observe that overall dual sourcing strategies \texttt{DualSrc-RL}, \texttt{TBS} are favorable in terms of rewards. Furthermore, \texttt{DualSrc-RL} is most profitable in all the run scenarios, and has $4\%$ reward gains over \texttt{TBS}.

\section{Capacity Management with Neural Coordination}
\label{sec: capacity}

% To do: 
% 1. add more explainations on model selection: why Neural Coordinator is a good method for constraining RL buy policies.
% 2. provide more technical details on Neural Coordinator.
% 3. call out in the paper that one of the key difference between the coordinator for our Dual Sourcing case and the one for their JIT case is that: our coordinator is aware of long lead orders, so can adjust its capacity costs to penalize future buys for the periods when it foresees that there is risk of violating capacity constraints.
% 4. List out Dagger algorithm for training buying policy and neural coordinator with multiple iterations.

Having formulated the unconstrained inventory control problem in $\mathbf{\mathcal{P}_1}$, now we consider a constrained situation, where network capacity constraints are introduced, and considered as part of the exogenous process. We are interested in studying the constrained problem because it is a challenging variant in real-world supply chain applications. A large retailer typically manages a supply chain for multiple products and has limited resources (such as storage) that are shared amongst all the products that retailer stocks. Specifically, we have the following set of formulas representing the storage capacity constraints,
\begin{align}
    G := \{ K_0, K_2, ..., K_{T-1} \}.
\end{align}

The constrained problem $\mathbf{\mathcal{P}_1}$ can be obtained by adding the capacity constraints \eqref{eq: storage_capacity_2} to $\mathbf{\mathcal{P}_1}$,

\begin{align}
     \mathbf{\mathcal{P}_2}: & ~\underset{\theta \in \Theta}{\max} ~\mathbb{E} \Big[\sum_{i \in \mathbb{A}} \sum_{t = 0}^{T-1} \gamma^{t} R_{t}^{i}(\theta) \Big], \\
    & \text{s.t.}   \nonumber \\
    % & I_{0}^{i} = \overline{I}^{i},  \label{eq: inv_init} \\
    % & (q^{i}_{J, t}, q^{i}_{L, t}) = \pi_{t}^{i} \big( \mathcal{H}_{t}, s_{t}^{i} ; \theta \big), \\
    % & \hat{\rho}^{i}_{J, t, 0}, \hat{\rho}^{i}_{J, t, 1}, \cdots \sim \mathrm{P}_{J} \big[{\rho}^{i}_{J, t, 0}, {\rho}^{i}_{J, t, 1}, \cdots | q^{i}_{J, t} \big], \label{eq: jit_qot_arrivals} \\
    %  & \hat{\rho}^{i}_{L, t+\delta_{L}} \sim \mathrm{P}_{L} \big[{\rho}^{i}_{L, t+\delta_{L}} | q^{i}_{L, t} \big], \label{eq: llt_qot_arrivals} \\
    % & I^i_{t_-} = I^i_{t-1} + \sum_{j=0}^{L_1} \min(U^{J,i}_{t-j},\tilde{q}^{J,i}_{t-j})\rho^{J,i}_{t-j,j} + \sum_{j=0}^{L_2} \min(U^{L,i}_{t-j},\tilde{q}^{L,i}_{t-j})\rho^{L,i}_{t-j,j}, \label{eq: i_t-calc_2}\\
    % & I_{t}^{i} = \max\{I_{t-}^{i} - d_{t}^{i}, 0 \}. \label{eq: onhand_periodend_2} \\
    & \sum_{i \in A} v^i  I_{t}^{i} \leq K_t,  \label{eq: storage_capacity_2}.
\end{align}

% \subsection{Penalized Dual Sourcing Problem}
% \label{sec:penalized_ds}
% Similar to \cite{eisenach2024ncc}, we construct the penalized Exo-IDP that is equivalent to applying Lagrangian relaxation to the constrained problem ($\mathbf{\mathcal{P}_2}$),

% \begin{align}
%      \underset{\lambda_t \leq 0}{\min} \;\; \underset{\theta \in \Theta}{\max} ~\mathbb{E} \Big[\sum_{i \in \mathbb{A}} \sum_{t = 0}^{T-1} \gamma^{t} R_{t}^{i}(\theta) + \sum_{t = 0}^{T-1} \lambda_t (\sum_{i \in A} v^i  I_{t}^{i} - K_t) \Big],
% \end{align}
% where $\lambda_t$ denotes the dual variables (also referred as capacity prices), $t \in [0, T-1]$. 

% This can be thought of as an $|A|+1$ agent problem, where there are $|A|$ product level buying policies, and one coordinator agent that sets capacity prices. We describe only the components that differ from the unconstrained problem.

For the reward function, we modify the unconstrained reward to incorporate a penalty according to the capacity prices,
\begin{align}
    R_{t}^{\lambda, i} \triangleq & \; R_{t}^{i} - \lambda_{t}  v^i  I_{t}^{i},
\end{align}
where $\lambda_{t}$ is the storage capacity price at time $t$.

\subsection{Forecasting the Capacity Prices by a Neural Coordinatator}
The role of coordinator agent is to predict capacity prices for the targeting inventory network given any capacity constraints as input. One example of a coordination mechanism is model predictive control (MPC), which would use forecasted demand to perform a dual cost search for the next $L$ periods. However, for an RL buying policy that uses many historical features, model predictive control would require forward simulating many features and it may be difficult to model the full joint distribution of all these features.

Instead, we apply a coordination approach as used in \citep{eisenach2024neuralcoordinationcapacitycontrol} by introducing a deep learning model for this problem. Specifically, we train a neural network to forecast the future prices of capacity that would be required to constrain the dual sourcing RL buying policy. Specifically, we learn a neural network to predict,
\begin{align}
    (\lambda_{t}, \lambda_{t+1}, ..., \lambda_{t+L}) = \phi_{\omega}(H_t, G),
\end{align}
where $H_t$ denotes the historical state vector and $G$ denotes the capacity constraints.

In the next, we introduce the optimization problem for learning the neural coordinator. First, assume a fixed dual sourcing buying policy $\theta$. Below we describe the ways in which the coordinator’s Exo-IDP deviates from the buying agent’s Exo-IDP.

The coordinator solves the following problem:
\begin{align}
     \mathbf{\mathcal{P}_3}: & ~\underset{\omega \in \Omega}{\min} ~\mathbb{E} \Big[ \sum_{t = 0}^{T-1} (\sum_{i \in A} v^i  I_{t}^{i} - K_t)_{+}^2 + || \lambda_t|| + \mathcal{L} \left( \lambda_t, H_{t}^{\lambda} \right) \Big], \label{eq: coordinator_loss}  \\
\end{align}
where $\mathcal{L} \left( \lambda_t, H_{t}^{\lambda} \right)$ denotes the total capacity price forecast error at time $t$,
\begin{equation}\label{eq:maxlikelihood}
\mathcal{L} \left( \lambda_t, H_{t}^{\lambda} \right)= \sum_{s=1}^{L} || \lambda_t - (\hat{\lambda}_{t-s}^{L})_{L-s} ||^2,
\end{equation}
which is the mean squared error (MSE) of all past forecasts for the current cost.
The training algorithm for the neural coordinator can be found in Appendix.
\subsection{Evaluation Results}
% To do:
% 1. remove EOM baselines
% 2. add DSRL+ MPC

In this section, we backtest our proposed dual sourcing buy policy with neural coordinator for storage capacity management. 

% \noindent \textbf{Dataset}
% For training both the buying and coordinating policies, we use a dataset of 250K fast products from the US marketplace from June 2017 to February 2020. Our out-of-sample-backtest period is from May 2022 to September 2023, again on a population of 250K fast products.

% \noindent \textbf{Capacity Constraint Paths}
% The constraint paths are from a space of functions of bounded variation. These are represented using the Haar wavelet basis, and we sample the coefficients in that basis from a multivariate Gaussian. In brief, the generated paths are scaled up proportional to demand and more details can be found in Section 4 of \cite{eisenach2024ncc}.
% % Figure  shows examples of the paths (they are scaled up proportional to demand).

\noindent \textbf{Compared Policies and Baselines}
We compare our unconstrained Dual Sourcing RL (\texttt{DualSrc-RL}) buy policy (as the baseline) with its two constrained variants. For the first variant, the neural coordinator is used to constrain the \texttt{DualSrc-RL} agent, and a MPC predictor is used to constrain \texttt{DualSrc-RL} as the second variant.

\noindent \textbf{Performance Metrics}
In addition to measuring reward achieved by the various policies, we consider several additional measures of constraint violation. They are (M1) mean constraint violation, (M2) mean violation on weeks where either unconstrained policy met at least 90\% of the limit, (M3) percent of weeks where the violation exceeded 10\% and (M4) percent of weeks where constraint violation exceeded 10\% and either unconstrained policy exceeded 90\% of the capacity limit.

\noindent \textbf{Results}
The Table \ref{tab: ds_rl_cap2_02} shows the results on the out-of-training backtesting period, where each combination of policy and coordinator were evaluated against 100 sampled storage constraint paths. Note that under all metrics (both violation and reward) the \texttt{DualSrc-RL} policy with neural coordinator outperforms \texttt{DualSrc-RL} with MPC. Although some of the violation metrics seem somewhat high, one should keep in mind that many of the capacity curves sampled will be highly constraining, much more so than in a real-world setting – in practice if the supply chain were so constrained, one would build more capacity.

\begin{table}[H]
 \vspace{-2mm}
 \caption{Out-of-distribution evaluation results.}
 \centering
 \footnotesize
 \begin{tabular}{@{}lrrrrrrrrrrrr@{}}
 \toprule
          & \multicolumn{2}{c}{} &  \multicolumn{4}{c}{Violations} & \\ \cmidrule(lr){4-7}      
Initialization       &   Buy Algo   &   Coordinator	 &	M1 	    &   M2         &   M3	   &	M4 &	Reward \\	\midrule
     	&	DualSrc-RL   &	  -	     &  28.9\%  &  54.7\%  &  35.8\%   &  67.8\%   &  \B{100}\\	
Onhand +     &  DualSrc-RL   &	Neural	 & \B{3.5 \%}& \B{6.7\%}& \B{9.1\%} & \B{17.3\%}&  99.7\\
Inflight    &	DualSrc-RL    &	  MPC      &    13.3\% &    25.1\%&     18.2\%    &   34.1\%    &    96.9 \\
            % &	BSHT    &	  -      &   19.7\% &    -    &   27.7\%    &   -       &  100.0\\
            % & 	BSHT    &	 MPC     &   4.9\% &   11.7\%   &   10.8\%   &   25.6\% &   99.9\\

  \bottomrule   
  \end{tabular}
\label{tab: ds_rl_cap2_02}
\end{table}

 The Figure \ref{fig:dsrl_cap2_01} below shows two examples of trajectories in the evaluation period for our \texttt{DualSrc-RL} (DS-RL) buy policies with their coordinator settings (unconstrained, Neural coordinator or MPC). We can see that the neural coordinator is able to constrain the on hand inventory within the capacity limit on the out-of-training backtesting period.

\begin{figure}[H]
  \centering
  \captionsetup[subfigure]{justification=centering}
  \begin{subfigure}{.5\linewidth}
    \centering
    \includegraphics[scale = 0.35]{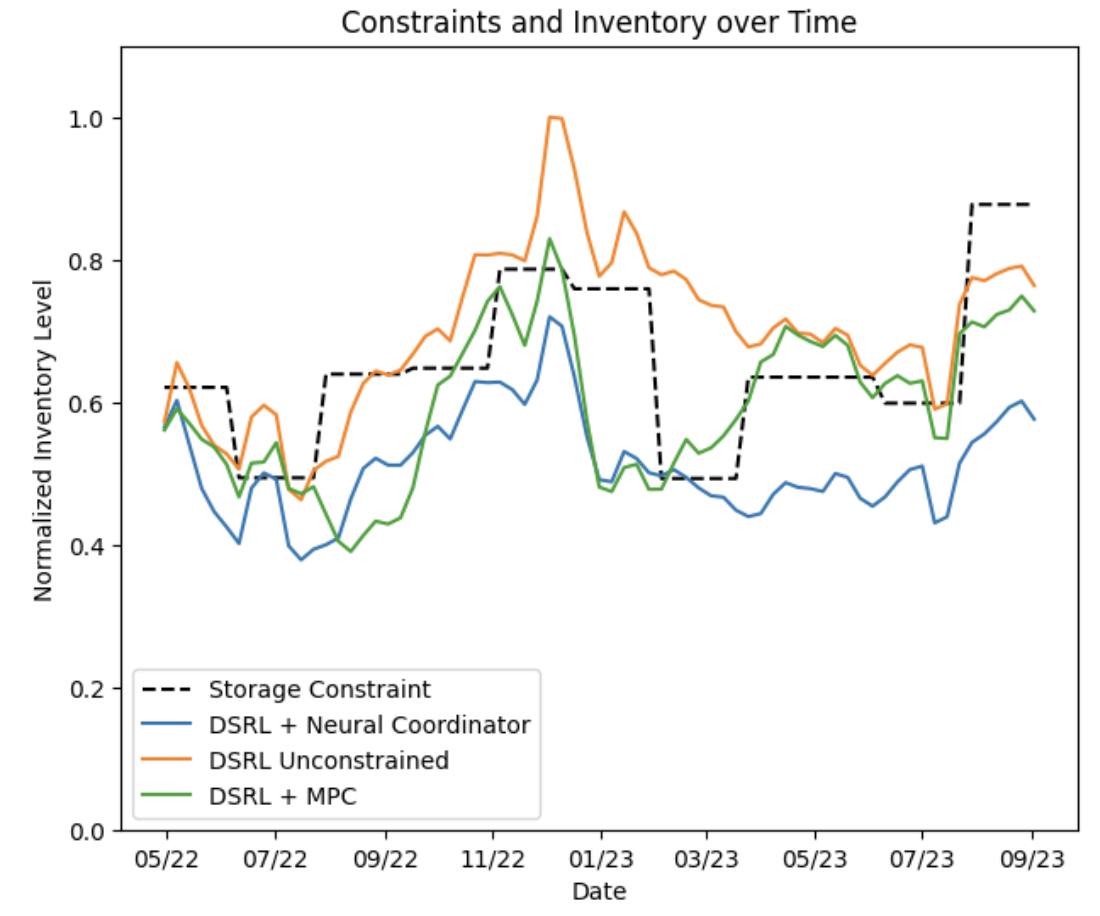}
    % \caption{\scriptsize Others (5) : \texttt{B001M20XA2} (Breakfast Bars).}\label{fig:anec same 1}
  \end{subfigure}\hfill
  \begin{subfigure}{.5\linewidth}
    \centering
    \includegraphics[scale = 0.35]{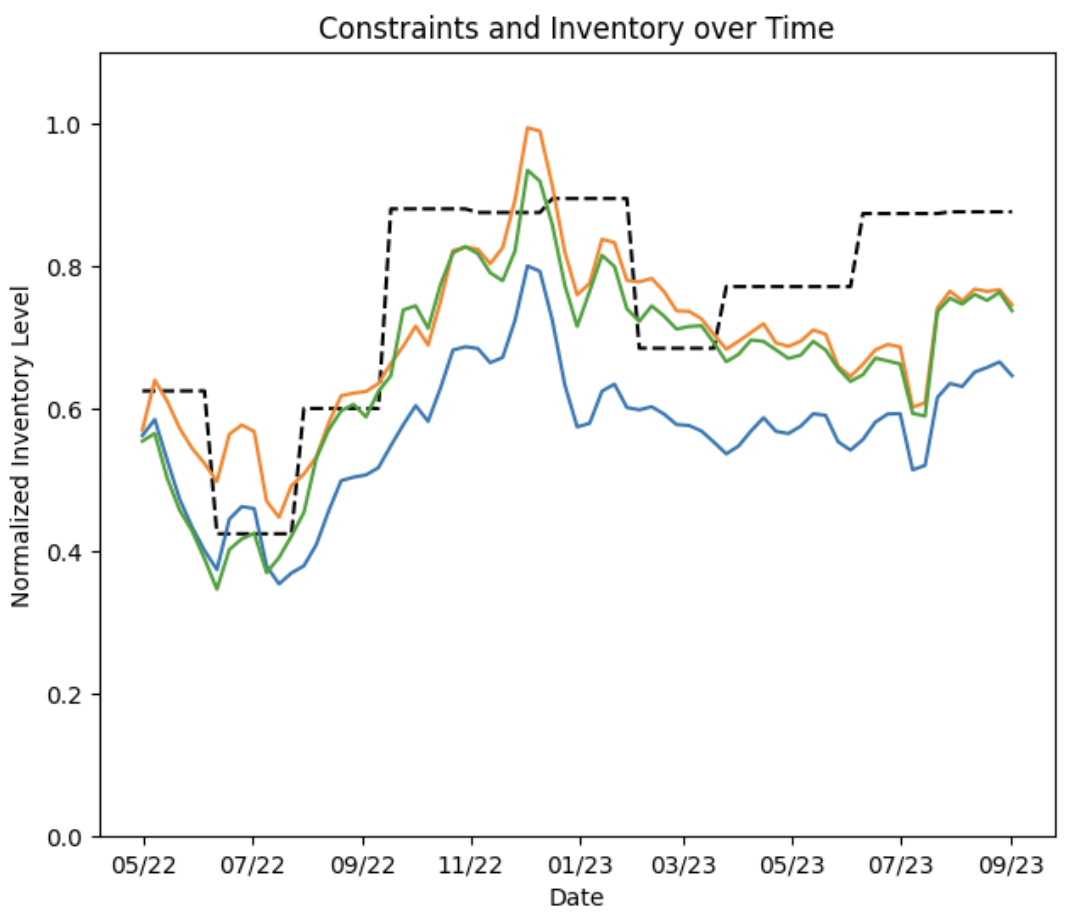}
    % \caption{\scriptsize  H-COGS<=10-S (1) : \texttt{B00NOU5562} (Clothing).}\label{fig:anec same 2}
  \end{subfigure}\hfill
  \caption{Inventory trajectories under different constraint paths during the out-of-training period.}
   \label{fig:dsrl_cap2_01}
  \end{figure}

\section{Conclusion and Future Work} \label{sec: conclusion}
% In this work, we explore dual sourcing inventory control methodologies that employ ordering decisions leveraging cost, lead time trade-offs between multiple sourcing channels. An Exo-IDP formulation is proposed for the dual sourcing problem, which facilitates development of simulators from historical datasets for training/backtesting of several policies. Consequently, we perform comprehensive ablation study experiments with RL based ordering policies vs competing OR baseline policies. Moreover, we also introduce capacity constraint modeling in our dual sourcing RL framework and apply a capacity control mechanism to generate capacity costs, guiding the system to adhere to target constraints. 

In this paper, we investigates ML-based inventory control methodologies with the consideration of stochastic supply chain processes that introduce scalable representation models to mitigate the supply risk and optimize the complex cross-product constraints/resources. Specifically, we investigate how to efficiently build a Deep RL framework for the problem by forecasting the real-world supply chain processes under a range of assumptions. We also introduce a capacity control mechanism, designed to forecast shadow prices given the capacity constraints of the inventory network. We highlight that instead of directly modeling the complex physical constraints into the learning pipeline and solving the problem as a whole, our approach breaks down those supply chain processes into different DL modules, leading to improved performance on larger real-world retail datasets. For future research, it is interesting to develop efficient modules for approximating other cross-product dependencies in real-world supply chain networks, e.g. containerization processes in global transportation, truck-load and placement. 

% We are considering the following directions for follow-on research:
% \begin{enumerate}
%     % \item In the current experimental setting, we still assume 100\% confirmation rate for LLT orders. Next, we will model the confirmation rate as a function of long lead time.
%     \item Depending upon the specific use case (e.g. Forward Buying vs Direct Import), we will fit QOT models on the appropriate populations for forecasting LLT arrivals and re-run backtests using data from the target population. For example, in the forward buying use case, a plausible model for arrivals is to first forecast the confirmed quantity, then the arrivals conditioned on the confirmation.
%     \item This work showed a proof of concept for capacity control with dual source RL. In practice, our target application may require constraining flow rather than storage (or both!).
%     \item A systematic investigation for properly capturing seasonal behavior at deeper levels of granularity.
%     \item Currently, we fix a pre-trained buying policy when training neural coordinator. But the optimality of the buying policy is also impacted by the underlying coordinator. To close this gap, one potential solution is to train the two models in a more collaborative way.
%     \item For future research, it is interesting to consider other cross-ASIN dependencies in Amazon's supply chain network, e.g. truck load constraints or containerization process in ocean transportation. 
% \end{enumerate}

\clearpage
\bibliographystyle{ims_nourl_eprint}
\bibliography{references}

\appendix
\clearpage

\section{Baseline Methods} \label{sec: baseline_methods}
\subsection{Improved Tailored Base Surge Policy}
We adopt a modified version of vanilla Tailored Base Surge (\texttt{TBS}) policy which has been described in \cite{xin2018asymptotic}. 
For each product $i$, TBS policy will place a \textit{dynamic} LLT order $q_{L, t}^{i, \texttt{TBS}}$ every period. In our problem setting, this LLT order will arrive with lead-time of $\delta_{L}$.
% as a consequence of Assumption \ref{assumption: llt fixed delay}.

Whereas, for orders through JIT channel, this TBS policy will first calculate a dynamic target order-upto-level $I^{i, \texttt{Tip}}_{t}$ via the production ``Horizon Tip Calculator'' method. Consequently, \texttt{TBS} policy places also places a dynamic order for the JIT channel i.e., $q_{J,t}^{i, \texttt{TBS}}$ that can bring back the inventory level to $I^{i, \texttt{Tip}}_{t}$. With on-hand inventory at the end of the period, $I_{t}^{i}$, the JIT order at time $t$ is given by:
\begin{align}
        q_{J,t}^{i, \texttt{TBS}} = \max \Bigg\{ 0, I^{i, \texttt{Tip}}_{t} - I_{t}^{i} - \Bigg[\sum_{\Tilde{t} = 0}^{t-1} \sum_{k = t}^{t + \delta_{ t}^{i,\texttt{Pred}}} \big( o^{J,i}_{\tilde{t},k - \tilde{t}}  +  o^{L,i}_{\tilde{t},k - \tilde{t}} \big) \Bigg] \Bigg\}, \label{eq: q_jit_tbs}
\end{align}
where $\delta_{t}^{i, \texttt{Pred}}$ is the median forecasted VLT at time $t$ for product $i$. In other words, Improved TBS subtracts on-hand and inflight inventory from the Horizon Tip to compute JIT orders for current period.
The order quantities $q_{J,t}^{i, \texttt{TBS}}, q_{L, t}^{i, \texttt{TBS}}$ will arrive according to the underlying quantity over time arrival models.

\noindent \textbf{Choice of LLT Order Input:} 
We set LLT order quantities $q_{L}^{i, \texttt{TBS}}$ as scaled 12-week rolling mean of product-level demands from training set, i.e., $q_{L,t}^{i, \texttt{TBS}} = \alpha \cdot \frac{1}{12} \cdot \sum_{\tilde{t}=t - 11}^{t - 1} d_{\tilde{t}}^{i}$. The order scaling factor $\alpha$ is used in our experiments as a search parameter for getting optimal \texttt{TBS} policy.

\subsection{Other Baseline Policies}
\begin{enumerate}
    \item We use a single source JIT Base Stock Policy with orders dictated by eq. \eqref{eq: q_jit_tbs} which we call \texttt{BaseStockHorizonTip} (BSHT).
    \item The existing RL buying policy for JIT single source from the literature \citep{madeka2022deep}.
\end{enumerate}

\section{Input Features} \label{sec: features}
\subsection{Featurization for Buying Policy} \label{sec: features_buy}
In terms of features, we mainly use the following feature list provided to the buying policy:
\begin{enumerate}
\item The current inventory level
\item Previous actions aiu that have been taken
\item Time series features
\subitem (a) Historical availability corrected demand
\subitem (b) Distance to public holidays
\subitem (c) Historical website glance views data
\item Static product features
\subitem (a) Product group
\subitem (b) Text-based features from the product description
\subitem (c) Brand
\subitem (d) Volume
\item Economics of the product - (price, cost etc.)
\item Capacity costs – past costs and forecasted future costs
\end{enumerate}

\subsection{Featurization for Neural Coordinator}  \label{sec: features_coordinator}
The neural coordinator takes the following aggregate/population level features:
\begin{enumerate}
\item Aggregated actions, inventory, demands for all current and previous times
\subitem (a) Order quantities
\subitem (b) Inventory
\subitem (c) Availability corrected demand
\subitem (d) Inbound
\subitem (e) All the above, but weighted by inbound and storage volumes

\item Forecasted quantities for next L weeks.
\subitem (a) Mean demand at lead time
\subitem (b) Inventory after expected drain at lead time
\subitem (c) All the above, but weighted by inbound and storage volumes

\item Other time series features
\subitem (a) Distance to public holidays
\subitem (b) Mean economics of products - (price, cost etc.), weighted by demand and volumes
\item Capacity costs (past costs and forecasted future costs)
\end{enumerate}

\section{Training Algorithms}
\label{sec_app: training}
\subsection{Training Algorithm for the Buy Policy}
Observe that constraints of $\mathbf{\mathcal{P}_1}$ \eqref{eq: p_1} are in fact definitions for different components of the reward function, and, therefore can be omitted from the optimization problem formulation otherwise. Next, we present the Direct Backpropagation (\texttt{DirectBP-DualSrc}) training algorithm for DRL policy.  

\begin{algorithm}
    \centering
\caption{Direct Backpropagation DRL training algorithm (\texttt{DualSrc-RL})} \label{alg:directbp}
\begin{algorithmic}
\State \textbf{Input:} set of products $\mathcal{A}$, training batch size $M$, step size: $\eta$, $\theta_{0} \in \Theta$.
\State \textbf{Initialize:} Batch Iterator: $b \leftarrow 1$.
\While{$\theta$ is not converged}
\State Sample mini-batch of products $\mathcal{A}_{b}$ of size $M$ from $\mathcal{A}$.
and set $R^{b} \leftarrow 0$.
\For{$i \in \mathcal{A}_{b}$}
\State $R^{i} \leftarrow 0$, $I_{0}^{i} \leftarrow \overline{I}^{i}$.
\For{$t = 0, \ldots, T^{\texttt{Train}} - 1$}
\State Place orders $\big(q^{i}_{J, t}, q^{i}_{L, t} \big) = \pi_{t}^{i}(H_{t}; \theta_{b-1})$.
\State Sample JIT, LLT arrivals $(o^{J,i}_{t,0},\dots, o^{J,i}_{t,L_1},o^{L,i}_{t,0},\dots, o^{L,i}_{t,L_2})$.
\State Update inventory $I_{t}^{i}$ according to \eqref{eq:arrivals} and \eqref{eqn:onhand_periodend1}. 
\State Collect reward $R_{t}^{i}$ according to \eqref{eqn:rewardfunc}
\State $R^{i} \leftarrow R^{i}_{t} + \gamma^{t} \cdot R_{t}^{i}$.
\EndFor
\State $R^{b} \leftarrow R^{b} + R^{i}$.
\EndFor
\State $\theta_{b} \leftarrow \theta_{b-1} + \eta \cdot \nabla_{\theta} \mathcal{P}_1^{b}\big|_{\theta = \theta_{b-1}}$.  // \textit{Update Parameters of Policy Network} $\pi(\cdot, \cdot; \theta)$. 
\State $b \leftarrow b + 1$.
\EndWhile
\end{algorithmic}
\end{algorithm}

\subsection{Training Algorithm for the Neural Coordinator}
We implement of the training algorithm for learning the neural coordinator for our dual sourcing problem. Specifically, we train the coordinator by solving the problem ($\mathbf{\mathcal{P}_3}$). Similar to the training algorithm for the dual sourcing buy policy, we apply the Direct Backpropagation algorithm to optimize the loss objective \ref{eq: coordinator_loss} in ($\mathbf{\mathcal{P}_3}$). The pseudo code of the training algorithm is shown in Algorithm \ref{alg:directbp_coordinator}.

\begin{algorithm}
    \centering
\caption{Training algorithm for the Neural Coordinator} \label{alg:directbp_coordinator}
\begin{algorithmic}
\State \textbf{Input:} set of products $\mathbb{A}$, training batch size $M$, step size: $\eta$, given buy policy $\theta$, \\ $\{\overline{I}^{i}\}_{i \in \mathbb{A}}, ~\delta_{L}, ~\epsilon$, initial neural coordinator $\omega_{0} \in \Omega$.
\State \textbf{Initialize:} Batch Iterator: $b \leftarrow 1$.
\While{stop criterion is not satisfied}
\State Sample mini-batch of products $\mathbb{A}_{b}$ of size $M$ from $\mathbb{A}$.
and set $R^{b} \leftarrow 0$.
\For{$i \in \mathbb{A}_{b}$}
\State $R^{i} \leftarrow 0$, $I_{0}^{i} \leftarrow \overline{I}^{i}$.
\For{$t = 0, \ldots, T^{\texttt{Train}} - 1$}
\State Collect instantaneous state and history vector $s_{t}^{i}, \mathcal{H}_{t}$.
\State Place orders $\Big(q^{i}_{J, t}, q^{i}_{L, t} \Big) = \pi_{t}^{i}(\mathcal{H}_{t}, s_{t}^{i} ; \theta)$.
\State Sample JIT, LLT arrivals $(o^{J,i}_{t,0},\dots, o^{J,i}_{t,L_1},o^{L,i}_{t,0},\dots, o^{L,i}_{t,L_2})$.
\State Collect reward $R_{t}^{i}$ and update inventory $I_{t}^{i}$. 
% \State $R^{i} \leftarrow R^{i}_{t} + (\gamma)^{t} \cdot R_{t}^{i}$.
\EndFor
% \State $R^{b} \leftarrow R^{b} + R^{i}$.
\EndFor

\State Update global state and compute coordination loss by Equation \ref{eq: coordinator_loss}.
\State $\omega_{b} \leftarrow \omega_{b-1} + \eta \cdot \nabla_{\omega} \mathcal{P}_3^{b}\big|_{\omega = \omega_{b-1}}$.  // 
\textit{Update Parameters of the Neural Coordinator} $\phi(\cdot, \cdot; \omega)$. 
\State $b \leftarrow b + 1$.
\EndWhile
\end{algorithmic}
\end{algorithm}
% Add the optimization problem
%% Add the training algorithm over the weekend!

\end{document}